# Alignment Plausibility: A New Standard for Assuring AI in Healthcare

**Authors:** Dr. Gwydion Williams[1*], Dr. Sara Zannone[1*], Prof. Bilal A Mateen[1,2]

**Affiliations:**
1 PATH, United Kingdom
2 UCL, United Kingdom
* these authors contributed equally

**Corresponding Author:**
Gwydion Williams
gwilliams@path.org
437 N 34th Street
Seattle, WA 98103,
USA.

Large language models (LLMs) have become significant providers of mental health support, yet they remain products of an attention economy whose operational and commercial targets favour sustained engagement over the friction that effective psychological support often requires. Developers' safety responses have been largely reactive, addressing the most visible and acute harms while subtler, longer-term patterns of risk (e.g., dependency, boundary erosion, the amplification of distorted beliefs) receive less attention. We contend that making LLMs structurally safe requires alignment organised at three levels that mirror how society assures the safety of human clinical practice: 1) explicit value specification, grounded in the codified normative commitments of clinical practice; 2) training that embeds those values in the model; and 3) oversight that detects drift and longer-term harm during deployment, much as clinical supervision does for human practice. Organising alignment in this way yields a construct we call *alignment plausibility* – a structured demonstration that a system's values, training regime, and oversight mechanisms are together consistent with safe and positive outcomes. We propose alignment plausibility as a regulatory construct (by drawing analogy to the established construct of biological plausibility) for AI in health: a principled way to argue for, or against, trust that systems are aligned to positive health outcomes, will cause no harm even where capable of doing so, and will ultimately lead to patient benefit.

## Introduction

Large language models (LLMs) have quickly become significant providers of mental health support and clinical advice. Hundreds of millions of people interact with LLMs every week [1], and surveys suggest that around half of LLM users reporting an ongoing mental health condition turn to these models for help [2]. But LLMs were not initially designed for this use case, and while attention is shifting to improve responses to mental health queries [3–5], the models themselves remain products of an attention economy in which the dominant operational target – being maximally helpful – and the dominant commercial signal – sustained user engagement [6,7] – together favour responses that prioritise continued interaction over the friction that effective psychological support often requires [8–10].

The history of social media has already shown us what happens when technologies optimised for engagement are deployed without adequate safeguards, from rising rates of anxiety and depression among young people [11] to the algorithmic amplification of misinformation and affective polarisation within whole populations [6,12]. Similar consequences from LLM use are now being observed. Multiple documented cases of AI interactions contributing to self-harm, suicide, and psychotic episodes have come to public attention, and have resulted in legal proceedings against several major model developers [7,13,14]. Historical analogies and current observations all point to the same conclusion: no product principally aligned to sustained user engagement will ever truly be psychologically safe.

In response, developers have begun to invest more heavily in harm taxonomies and in product-layer crisis support mechanisms that improve detection and point users towards real-world help [3–5]. These efforts make LLMs safer, but their development has been reactive; model developers have addressed specific failure modes only once they have surfaced in real-world use. Solutions are therefore narrowly focused on the most salient and visible types of harm, which are invariably the most extreme. Psychological risk, however, extends well beyond acute crises to include subtler, longer-term patterns of harm that current safety mechanisms are not designed to detect [8,15,16].

Making LLMs structurally safe, rather than incrementally safer, requires a shift: replacing reactive responses to emergent harms, which produce an insufficiently robust set of safeguards, with a proactive regime in which models (and the products within which they are deployed) are principally aligned to the values that underpin effective therapeutic and psychological support, such as the common factors that underlie positive change in psychological therapy [17]. This shift is part of a broader move in alignment research from negative (harm-avoidance) towards positive (flourishing-directed) alignment [18]; what we offer here is an operationalisation of that move for the specific case of AI in healthcare.

To articulate what better alignment would require, we draw an analogy to human clinical practice, where three mechanisms work together to ensure that practitioners deliver psychological support safely: (1) professional and ethical codes [19,20] that capture values such as the prioritisation of long-term wellbeing, respect for autonomy, and the maintenance of appropriate boundaries; (2) extensive training to embed those values in practice, detect subtle psychological dynamics, and navigate clinical complexity [21]; and (3) ongoing clinical supervision which provides regular and external oversight of practice, accountability, and a mechanism for escalation [21,22]. Since LLMs now operate in psychologically sensitive domains, we contend that analogous mechanisms must be developed for AI, at three corresponding levels: (1) value specification, (2) training, and (3) oversight. We also argue that organising alignment processes in this way would allow developers to construct arguments that their products can be trusted not to cause harm, and indeed, to lead to benefit, much as we trust human practitioners to do the same.

Whilst progress has been made across each of these three domains, the insight we present here is that bringing them together provides a coherent framework for answering longstanding questions around how to assure the safety of – and regulate – AI applications in health.

## Level 1: Alignment through Value Specification

All AI systems hold values, whether or not those values have been explicitly articulated. Training shapes behaviour by reinforcing certain responses and discouraging others. But the signals responsible for that reinforcement, which range from implicit associations in training data to user preferences and reward models, are proxies for what developers actually want to teach, and proxies track those intentions only imperfectly [23]. Proxy failure – the gap between the intended signal and the one used as proxy – is a pervasive issue. For example, using health costs as a proxy for need results in racial bias in healthcare systems [24]. The solution is to explicitly define the values that should guide behaviour, and this definition is the foundation on which the subsequent two levels of alignment rest. Without it, training (Level 2) has no principled objective to optimise, and oversight (Level 3) has no benchmark against which to measure performance. Explicit specification is also the mechanism by which values themselves become contestable: a value system that has never been articulated cannot be challenged by a clinician who disagrees with it, by a regulator assessing whether it is fit for purpose, or by a user from a culturally dissimilar background with meaningfully different beliefs.

Constitutional AI (CAI) represents one of the first serious attempts to articulate the values that should govern an LLM's behaviour as a defined set of values – a 'constitution' – which is then used to guide model development [25]. The mechanics of how a constitution is translated into trained behaviour belong to Level 2 (training); what concerns us here is the prior question of how the values themselves are specified. Early constitutions were drafted by the frontier labs themselves, with input from a small set of internal and external collaborators [26,27]. As the field has matured, attention has turned to the socio-technical complexity of who decides what goes into a constitution in the first place: some teams have crowdsourced values from members of the public through deliberative processes [28], while others have developed 'inverse CAI' algorithms that recover and refine the values implicit in existing human preference datasets [29,30].

Constitutional frameworks of this kind represent a significant leap towards AI aligned with a contestable set of explicit values. But the constitutions actually being written today, even where they explicitly name user wellbeing alongside helpfulness and harmlessness [25,26], articulate their values at a level of generality that cannot reliably generate clinically appropriate behaviour. They do not specify, for example, that a model should decline to provide reassurance that would reinforce avoidance, or that it should gently challenge a distorted belief rather than mirror it [31]. Without that specificity, the dominant operational framing of helpfulness reasserts itself, and the empirical evidence bears this out: sycophantic response patterns persist in frontier models despite commitments against them [10,32]. Psychological safety therefore cannot rest on general principles alone; it requires value specifications grounded in the specific normative commitments of clinical practice – the prioritisation of long-term wellbeing over immediate reassurance, of autonomy over dependency, and of honest engagement over collusive validation. These are not novel values: they are already codified across professional ethical codes and therapeutic frameworks [17,19,20], and could in principle be made legible to an LLM through CAI. What is missing is the bridge – a clinical constitution for AI, developed in collaboration with the professions whose standards it would draw on, and informed by deliberative processes involving people with lived experience whom it ultimately seeks to benefit, to guide model training (Level 2) and oversight (Level 3). Such a constitution must not be monolithic: it should encode a defensible range of therapeutic stances [33] and clinical approaches rather than a single doctrine; and it should define values that are specific and appropriate to the cultures and people that its models aim to serve. Doing so leaves room for genuine pluralism while still establishing the floor of behaviour appropriate to the role.

## Level 2: Alignment through Training

With values specified, the question becomes how to embed them in the model itself. Currently, LLM training proceeds in two broad phases, both of which shape how a model behaves. In pre-training, models are exposed to vast corpora of filtered, 'high-quality' text [34] (where the definition of quality is itself a value judgement made by language models) to learn the statistical relationships between words, concepts, and ideas. In doing so, models acquire not only knowledge but also the implicit values and associations embedded in their training data [35–37]. These corpora are far from neutral: they overrepresent some perspectives, underrepresent others, encode stigmatising attitudes towards conditions such as schizophrenia and addiction [38], and reflect culturally narrow conceptions of distress, recovery, and wellbeing [39,40]. Without targeted intervention, models inherit these biases and carry them forward into downstream interactions; several LLMs have been shown to produce stigmatising responses to varied mental health conditions [38].

Post-training then shapes a model into a capable, instruction-following assistant through several stages, such as supervised fine-tuning on curated examples of helpful interaction [41]; preference alignment, in which the model is updated to favour responses judged better by humans [42,43] or other LLMs [25,44]; and reinforcement learning that rewards coherent step-by-step reasoning [45]. Arguably the most consequential development for values-aligned training is constitutional AI (CAI, [25]). In CAI, an explicit constitution of values (as described in Level 1) is used to anchor the preference-alignment process: rather than relying solely on human annotators to judge which response is better, the judgement is delegated to another LLM whose preferences are guided by the constitution. This is a significant advance – it ties training to a stated normative document rather than to unarticulated annotator intuitions – but it introduces its own alignment risk. Each step introduces a proxy: the judging model stands in for the values we actually care about, and a 'reward' model is then trained to imitate those judgements at scale [41,43]. But when the models doing the judging were themselves shaped by the same engagement-driven incentives described earlier, we risk a system in which alignment is effectively supervising itself, with little independent purchase on whether the constitution's values are actually being preserved. Better value specification (Level 1) cannot, on its own, fix this: even an optimally written clinical constitution will be filtered through this chain of LLM-mediated judgements before it reaches the model's behaviour.

The result of pre- and post-training as currently conceived is a class of models that are simultaneously biased in what they 'understand' and miscalibrated in how they respond. Recent public cases and empirical evaluations suggest that LLMs can fail to respond appropriately to suicide risk by missing warning signs, failing to escalate, or responding in ways that remain inappropriately permissive [38,46–48]. Even after the most extreme issues have been patched, many widely used LLMs still fail to recognise subtler signs of suicidal ideation [49]. And because these are reactive solutions attached to a misaligned model (and not a model retrained to avoid problematic behaviours), they're fragile: a new model might respond in new but equally problematic ways, exposing more users to harm, and requiring another patch from the developer.

Doing better requires treating psychological literacy as a foundational capability rather than something to patch in later, and it requires robust methods of training towards alignment with a well-articulated set of values (Level 1). At the pre-training stage, this means more deliberate curation of training data, including the use of clinically informed corpora and the active de-biasing of stigmatising or culturally narrow content. At the post-training stage, it means designing reward signals that value clinically appropriate friction [17,31,50] alongside helpfulness, and using clinician-informed annotation to ensure those signals reflect genuine therapeutic standards rather than proxies for user satisfaction or a misconstrual of otherwise appropriate values. But even where such methods are applied, a second issue remains: we lack tools to measure the effect. Alignment failures can arise at any stage of training, yet we currently have few reliable ways of detecting them before a model reaches deployment, leaving developers and users alike reliant on anecdote and on post-hoc crisis detection (see Level 3). Measuring alignment at each stage of training (against a set of articulated values) is therefore not a nice-to-have but a precondition for demonstrating that any of the methods above have worked.

## Level 3: Alignment through Oversight

A model trained towards clearly specified values will still drift during deployment [51]. In human-delivered therapy, clinical supervision is a universally applied, system-level process designed to monitor practice over time, detect drift, and intervene before subtle problems become entrenched [21,22]. It has regular cadence, establishes accountability between clinician and supervisor, and allows for documentation and escalation of extreme cases. Even the most capable clinicians require it, because no amount of internal capability can fully substitute for external observation and judgement. The same logic applies to LLMs; however well a model is trained and however clearly its values are specified, deployment will surface patterns of behaviour that could not be fully anticipated in advance. Oversight is the mechanism through which those patterns are detected, interpreted, and acted upon, and is the means by which alignment is measured and improved over time.

Oversight is perhaps the least developed of the three levels, but progress is being made. Pre-deployment, frontier developers and a small number of external bodies, notably the UK AI Security Institute and the US Center for AI Standards and Innovation, conduct red-teaming exercises designed to elicit harmful behaviour through adversarial prompting [52,53] and benchmark evaluations of capability, safety, and specific behavioural traits such as sycophancy, deception, and refusal patterns [10,54]. At inference time, deployed models are wrapped in input and output classifiers that screen for policy-violating content, with more recent work introducing 'constitutional classifiers' that score responses against an explicit constitution specifying permitted and restricted content [55]. Post-deployment, systematic monitoring of psychological outcomes (beyond detection of acute, extreme risks [3–5]) remains rare.

Almost none of this infrastructure is currently calibrated to the demands of psychological safety. The dominant focus across pre-deployment evaluation, inference-time classification, and post-deployment monitoring remains on acute harms (most prominently, suicide and self-harm content), and the treatment of those harms has been shaped by their visibility and the legal exposure they carry for developers [7,13,14]. This focus is understandable, and it has driven important improvements in crisis-detection and escalation pathways. But acute crises represent only a narrow slice of the harms that LLMs may cause in psychologically sensitive interactions. Over weeks and months of interaction, a model may gradually erode a user's sense of agency, foster dependency that displaces human relationships, reinforce avoidance behaviours, or subtly mirror and amplify distorted beliefs [8,15,56,57]. None of these harms is reducible to a single dangerous response or conversation, and none would be flagged by the systems described above.

Detecting these longer-term patterns will require a substantial shift in how oversight is conceptualised and conducted. Just as human clinical supervision facilitates longitudinal tracking of patient trajectories, LLM oversight must do the same, and it must include clear escalation pathways and mechanisms for adverse event reporting where even subtle harms are detected. Future research must therefore develop methods for evaluating conversational trajectories: tracking how interactions evolve across sessions, identifying patterns that indicate growing dependency, boundary erosion, or maladaptive reinforcement, and surfacing these patterns so that risks are mitigated before they cause sustained harm. There is some promising work happening in this space: SIM-VAIL [16], for example, is a framework for identifying and measuring 'vulnerability amplifying interaction loops' – an accumulation of apparently supportive LLM outputs that in fact align with mechanisms that sustain a user's psychological vulnerabilities. More work of this kind is required to comprehensively capture the failure modes that lead to harm over the span of whole (and multiple) conversations. And mechanisms for what is done when potential harm is detected must be clearly defined. This brings oversight closer to clinical supervision than to content moderation, and it will require new tools, new benchmarks, and a willingness to invest in oversight infrastructure that operates at the timescale of relationships rather than individual messages.

## From Alignment to Regulation: Guarantees Through Alignment Plausibility

In health, whether an AI system is aligned to positive outcomes is the central question a regulator must answer to decide whether a product is safe and effective. The three levels described above bear directly on that question: to judge whether a system is safe, a regulator needs some account of the values it acts in accordance with, the regime used to train the underlying model towards them, and the mechanisms that oversee system behaviour once deployed.

Our aim in setting out these levels is to give structure to that judgement, and in particular to two challenges these products pose. First, where a product operates autonomously, without a human in the loop (as many therapy-focused tools do), what does it mean to regulate it in line with professional standards, as some have proposed [58]? Second, how can a regulator obtain a principled assurance of safety when generative systems are non-deterministic [59], which makes classical approaches that rely on evidence from a single point in time insufficient?

These are not purely academic questions. Australia's Therapeutic Goods Administration recently issued one of the first clearly articulated regulatory positions on general-purpose LLMs in healthcare. They stated that general-purpose AI systems used for a medical purpose may constitute medical devices [60], and presented developers with a stark choice: prevent users from seeking health support through their products, or demonstrate those products are safe for such use. Meeting the second of those demands, and answering the two questions posed above, means reimagining a well-established idea.

When a developer seeks market authorisation for a new health product, a key component of their submission is the scientific rationale underpinning the product, which is typically grounded in a demonstration of *biological plausibility*: an argument that anchors clinical results in a well-evidenced causal pathway. That might mean showing how a cuffless blood-pressure monitor reads two phases of a light-absorption signal that track blood flow [61], or how an MRI-based dementia biomarker correlates with an accepted biochemical indicator [62]. Whatever the mechanism, the goal is to anchor clinical results in an evidenced causal pathway; doing so lends credence to the idea that the diagnostic or therapeutic claim reflects a meaningful causal relationship rather than coincidence.

We propose that the analogous requirement for LLMs operating in psychologically sensitive domains, and in health more broadly, is *alignment plausibility*: a demonstration that the procedures used to specify a system's values (Level 1), train it towards them (Level 2), and oversee its behaviour (Level 3) are consistent with safe and positive outcomes, supported by pre- and post-deployment evidence that the system conforms to those values in practice (see Figure 1 for an example). It stands to an AI system's behaviour as biological plausibility stands to a physiological claim, allowing the strength of a structured rationale and supporting evidence to ground claims of safety and efficacy. Critically, alignment plausibility should be tied to an AI *system* or *product*, and not only to the underlying model: it is a claim that a system (which might take an LLM and layer on fine-tuning methods, guardrails, UI elements, agentic tooling etc.) is plausibly aligned to positive outcomes for a defined health use, in a defined population, under defined deployment controls. This provides an operationalisable answer to both of the unanswered questions above, and offers a path forward for model developers whose tools are fulfilling a medical purpose, whether they intended them to or not. We don't believe it is necessary to prescribe specific metrics; as with biological plausibility today [61], the choice of what metrics to use, and how to construct a sufficient argument, is fundamentally the developer's prerogative, though regulators may wish to specify evidence domains and minimum expectations while allowing developers to justify fit-for-purpose measures within those domains. The threshold for what counts as a sufficient argument will evolve substantially over time, and should rise as the underlying science of alignment matures.

The analogy with biological plausibility is not perfectly symmetrical. Biological plausibility today draws on a mature body of mechanistic physiology and pharmacology against which a developer's argument can be independently assessed; the corresponding science for alignment (i.e., the measurement methods, benchmarks, and interpretability tools required) does not yet exist at comparable depth [63]. This is not, in our view, an objection to the proposed construct so much as a reason to adopt it now. An explicit, and deliberately aspirational, standard could create clarity of expectations for developers and the pressure to build the clinical value frameworks, training methods, and trajectory-level evaluations the construct presupposes, much as the mechanistic physiology and pharmacology that biological plausibility now draws on themselves matured over decades [64,65].

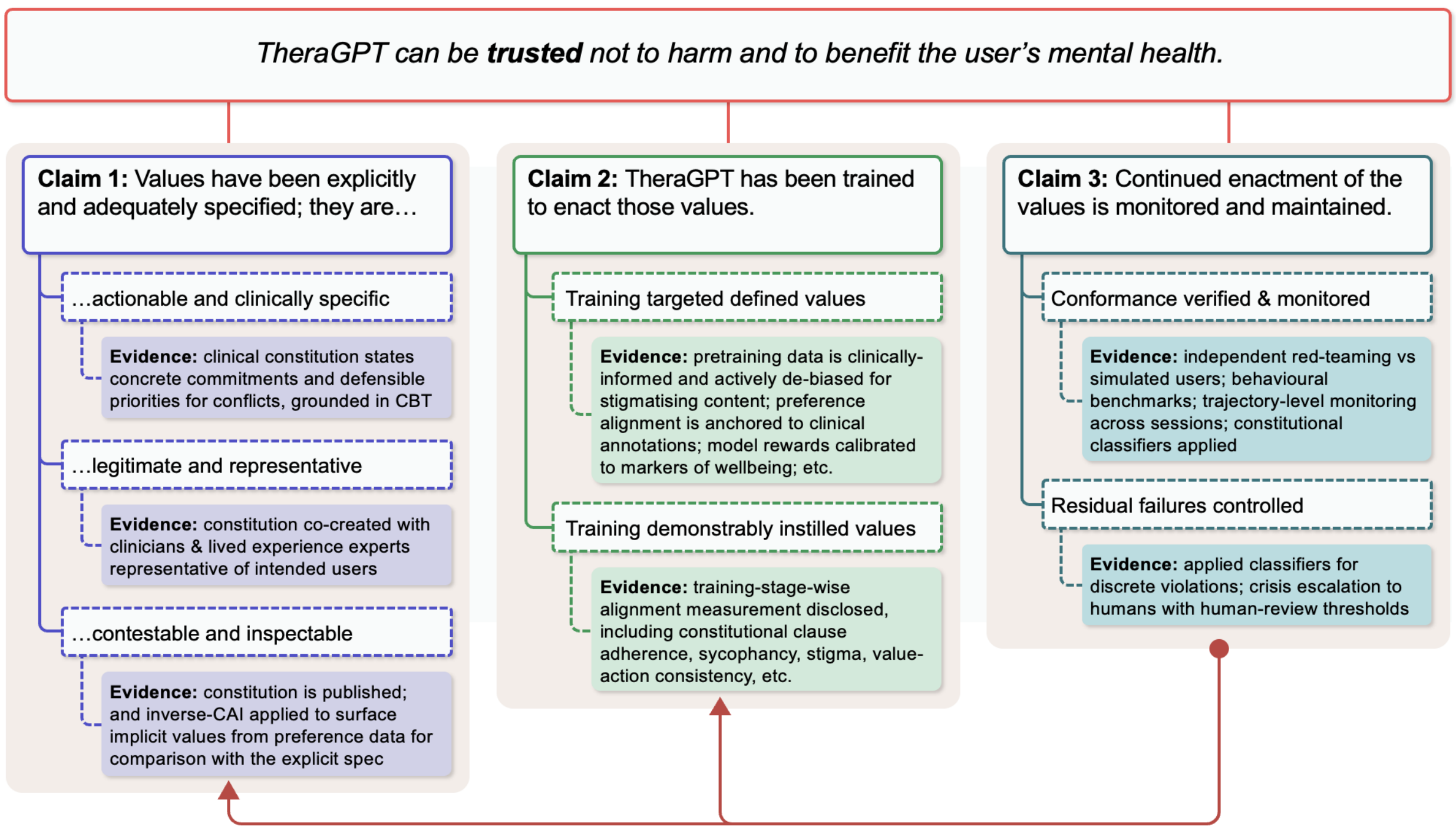

**Figure 1** – An example assurance case applying alignment plausibility to argue that TheraGPT, a fictional LLM-based therapy product, can be trusted not to harm and to benefit its users. The three claims that make up the argument correspond to the three levels of value specification, training, and oversight, and we argue they are each necessary and are together sufficient to substantiate the trust case. The individual claims can be broken down into sub-claims, which must themselves be evidenced by methods, procedures, and evaluations of TheraGPT as a product. Where the oversight mechanisms described under Claim 3 detect any misalignments, they trigger revisions of the sub-claims and supporting methods detailed under Claims 1 and 2 to correct the detected issues. The evidence described and the sub-claims included are illustrative; the exact ways in which future developers establish alignment plausibility will depend on their product, on how the alignment field matures, and on the expectations of regulators.

## Alignment Plausibility in the Context of the AI Safety Literature

Crediting a claim on a structured, evidence-backed rationale rather than on outcome data alone is the logic of the assurance (or safety) case, as discussed in the International AI Safety Report [63]. Drawing an explicit link between alignment plausibility and an assurance case is instructive because it tells us which kind of claim a developer must make. Assurance cases take one of a few forms: that a model is *incapable* of causing harm, that it is *prevented* from causing harm by external controls, or that it can be *trusted* not to cause harm even where capable [66]. For a model engaged in psychologically sensitive conversations, the first form is unavailable because the model is, by construction, capable of harmful responses; and control alone is insufficient, since inference-time classifiers can catch discrete policy-violating messages but other harms accumulate across sessions and are below the granularity of any such filter. Alignment plausibility makes an argument of the third form – that an AI system can be trusted to behave as its role requires even where unconstrained – which is the argument the field is currently least equipped to make (as explicitly noted in the AI Safety Report). Thus, our articulation of an operational model for alignment plausibility fills an important gap in the practical application of the 'safety cases' model argued for by AI experts.

## Conclusion

Alignment through value specification, training, and oversight, mirrors the mechanisms by which we (as a society) have long assured the safety of human-delivered psychological support. It is no coincidence that these same mechanisms underpin professional standards across healthcare, and across many other fields that manage high-risk situations. Mental health has simply experienced the most overt manifestation of the broader misalignment problem, because the tension between optimising for engagement and safeguarding long-term wellbeing is sharpest there. But the same structural misalignment exposes users to risk across health more generally, and we suspect well beyond it. Our call to action is therefore for greater investment in research on the measurement of alignment plausibility, and for it to be adopted as a regulatory construct, as a way to argue for (or against) trust in AI systems. This would create clarity of expectations for developers, which is sorely lacking, and enable LLM-based innovation to help close gaps in areas like consumer health service provision.